\documentclass[nohyperref]{article}

\usepackage{microtype}
\usepackage{graphicx}
\usepackage{subfigure}
\usepackage{booktabs} % for professional tables
\usepackage{multicol}
\usepackage{hyperref}

\usepackage[accepted]{icml2022}

\usepackage{amsmath}
\usepackage{amssymb}
\usepackage{mathtools}
\usepackage{amsthm}
\usepackage{graphicx}

\usepackage{hyperref}       % hyperlinks
\usepackage{url}            % simple URL typesetting
\usepackage{booktabs}       % professional-quality tables
\usepackage{amsfonts}       % blackboard math symbols
\usepackage{nicefrac}       % compact symbols for 1/2, etc.
\usepackage{microtype}      % microtypography
\usepackage{xcolor}         % colors
\usepackage{tikz}

\usepackage{float}
\usetikzlibrary{arrows.meta}
\usetikzlibrary{calc}
\usepackage{caption}

\usepackage{comment}
\usepackage{authblk}
\usepackage{cancel}

\makeatletter
\renewcommand\AB@affilsepx{, \protect\Affilfont}
\makeatother

\usepackage{xcolor}
\colorlet{linkcolor}{blue!70!black}
\hypersetup{
  colorlinks,
  linkcolor={red!50!black},
  citecolor={blue!50!black},
  urlcolor={blue!80!black}
}

\graphicspath{{images/}}

\usepackage[capitalize,noabbrev]{cleveref}

\theoremstyle{plain}
\newtheorem{theorem}{Theorem}[section]
\newtheorem{proposition}[theorem]{Proposition}

\theoremstyle{definition}

\theoremstyle{remark}

\usepackage{graphicx}
\usepackage{color}

\usepackage[ruled,vlined,linesnumbered]{algorithm2e}
\usepackage{algorithmic}
\usepackage{stmaryrd}
\usepackage[inline]{enumitem}
\usepackage{url}

\usepackage{tikz}
\usetikzlibrary{calc}

\usepackage{pgfplots}
\usepackage{xcolor}
\usepackage{bbm}
\usepackage{ifthen}
\usepackage{xargs}

\newtheorem{assumptionF}{\textbf{F}\hspace{-3pt}}
\crefformat{assumptionF}{{\textbf{F}}#2#1#3}

\Crefname{assumptionB}{\textbf{B}\hspace{-3pt}}{\textbf{B}\hspace{-3pt}}
\crefname{assumptionB}{\textbf{B}}{\textbf{B}}

\Crefname{assumptionC}{\textbf{C}\hspace{-3pt}}{\textbf{C}\hspace{-3pt}}
\crefname{assumptionC}{\textbf{C}}{\textbf{C}}

\Crefname{assumptionH}{\textbf{H}\hspace{-3pt}}{\textbf{H}\hspace{-3pt}}
\crefname{assumptionH}{\textbf{H}}{\textbf{H}}

\Crefname{assumptionT}{\textbf{T}\hspace{-3pt}}{\textbf{T}\hspace{-3pt}}
\crefname{assumptionT}{\textbf{T}}{\textbf{T}}

\Crefname{assumptionT}{\textbf{T}\hspace{-3pt}}{\textbf{T}\hspace{-3pt}}
\crefname{assumptionT}{\textbf{T}}{\textbf{T}}

\Crefname{assumptionL}{\textbf{L}\hspace{-3pt}}{\textbf{L}\hspace{-3pt}}
\crefname{assumptionL}{\textbf{L}}{\textbf{L}}

\Crefname{assumptionQ}{\textbf{Q}\hspace{-3pt}}{\textbf{Q}\hspace{-3pt}}
\crefname{assumptionQ}{\textbf{Q}}{\textbf{Q}}

\Crefname{assumptionAR}{\textbf{AR}\hspace{-3pt}}{\textbf{AR}\hspace{-3pt}}
\crefname{assumptionAR}{\textbf{AR}}{\textbf{AR}}

\newcommand\diaW{14}
\newcommand\diaSW{6}
\newcommand\diaH{5} %\newcommand\diaH{9}
\newcommand\diaJump{2.75}
\newcommand\nextRow{1.25}

\newcommand\SnextRow{1.5}

\newcommand\imW{0.07}
\newcommand\imSW{0.3}

\newcommand\imOp{0.6}
\newcommand\bend{5}

\usepackage{bm}
\usepackage{wrapfig}

\newcommand{\dive}{\mathrm{div}}

\newcommand{\XM}{\mathcal{X}(\mathcal{M})}

\def\hlf{\hat{\ell}^f}
\def\hlb{\hat{\ell}^b}

\def\pdata{p_{\textup{data}}}

\def\pprior{p_{\textup{prior}}}

\def\Pens{\mathcal{P}}

\newcommandx{\expec}[2]{{\mathbb E}\left[#1 \middle \vert #2  \right]} %%%% esperance conditionnelle

\newcommandx{\norm}[2][1=]{\ifthenelse{\equal{#1}{}}{\left\Vert #2 \right\Vert}{\left\Vert #2 \right\Vert^{#1}}}
\newcommandx{\normLigne}[2][1=]{\ifthenelse{\equal{#1}{}}{\Vert #2 \Vert}{\Vert #2\Vert^{#1}}}

\def\bfc{\mathbf{c}}
\def\bfY{\mathbf{Y}}

\def\bfX{\mathbf{X}}
\def\bfhX{\hat{\mathbf{X}}}
\def\bfW{\mathbf{W}}

\def\bfZ{\mathbf{Z}}

\def\bfZ{\mathbf{Z}}

\def\bfB{\mathbf{B}}

\def\Qbb{\mathbb{Q}}

\def\Pbb{\mathbb{P}}

\def\rset{\mathbb{R}}

\def\nset{\mathbb{N}}

\def\rmd{\mathrm{d}}

\def\rmc{\mathrm{C}}

\newcommand{\M}{\mathcal M}

\newcommandx{\functionspace}[2][1=+]{\mathbb{F}_{#1}(#2)}
\newcommand{\argmin}{\operatorname*{arg\,min}}

\newcommandx{\VarDeux}[3][3=]{\operatorname{Var}^{#3}_{#1}\left\{#2 \right\}}

\newcommand{\LeftEqNo}{\let\veqno\@@leqno}

\newcommand{\N}{\ensuremath{\mathbb{N}}}

\newcommand{\PE}{\mathbb{E}}

\newcommandx{\Vnorm}[2][1=V]{\| #2 \|_{#1}}
\newcommandx{\VnormEq}[2][1=V]{\left\| #2 \right\|_{#1}}
\newcommandx\probaMarkovTilde[2][2=]
{\ifthenelse{\equal{#2}{}}{{\widetilde{\mathbb{P}}_{#1}}}{\widetilde{\mathbb{P}}_{#1}\left[ #2\right]}}

\newcommand{\expeLigne}[1]{\PE [ #1 ]}

\def\ie{i.e.}

\def\eqsp{\;}

\newcommand{\ccint}[1]{\left[#1\right]}

\newcommandx{\weight}[2][2=n]{\omega_{#1,#2}^N}

\newcommandx\sequence[3][2=,3=]
{\ifthenelse{\equal{#3}{}}{\ensuremath{\{ #1_{#2}\}}}{\ensuremath{\{ #1_{#2}, \eqsp #2 \in #3 \}}}}

\newcommandx\sequenceD[3][2=,3=]
{\ifthenelse{\equal{#3}{}}{\ensuremath{\{ #1_{#2}\}}}{\ensuremath{( #1)_{ #2 \in #3} }}}

\newcommandx{\sequencen}[2][2=n\in\N]{\ensuremath{\{ #1_n, \eqsp #2 \}}}
\newcommandx\sequenceDouble[4][3=,4=]
{\ifthenelse{\equal{#3}{}}{\ensuremath{\{ (#1_{#3},#2_{#3}) \}}}{\ensuremath{\{  (#1_{#3},#2_{#3}), \eqsp #3 \in #4 \}}}}
\newcommandx{\sequencenDouble}[3][3=n\in\N]{\ensuremath{\{ (#1_{n},#2_{n}), \eqsp #3 \}}}

\newcommand{\opnorm}[1]{{\left\vert\kern-0.25ex\left\vert\kern-0.25ex\left\vert #1
    \right\vert\kern-0.25ex\right\vert\kern-0.25ex\right\vert}}

\newcommandx{\CPE}[3][1=]{{\mathbb E}_{#1}\left[#2 \middle \vert #3  \right]} %%%% esperance conditionnelle
\newcommandx{\CPELigne}[3][1=]{{\mathbb E}_{#1}[#2  \vert #3  ]} %%%% esperance conditionnelle
\newcommandx{\CPEsq}[3][1=]{{\mathbb{E}^{1/2}}_{#1}\left[#2 \middle \vert #3  \right]} %%%% esperance conditionnelle
\newcommandx{\CPVar}[3][1=]{\mathrm{Var}^{#3}_{#1}\left\{ #2 \right\}}
\newcommand{\CPP}[3][]
{\ifthenelse{\equal{#1}{}}{{\mathbb P}\left(\left. #2 \, \right| #3 \right)}{{\mathbb P}_{#1}\left(\left. #2 \, \right | #3 \right)}}

\newcommandx{\osc}[2][1=]{\mathrm{osc}_{#1}(#2)}

\newcommand{\ensembleLigne}[2]{\{#1\,:\eqsp #2\}}

\newcommand\coupling[2]{\Gamma(\mu,\nu)}

\newcommandx{\KL}[2]{\operatorname{KL}\left( #1 | #2 \right)}
\newcommandx{\KLsqrt}[2]{\operatorname{KL}^{1/2}\left( #1 | #2 \right)}
\newcommandx{\Jef}[2]{\operatorname{J}\left( #1 , #2 \right)}
\newcommandx{\JefLigne}[2]{\operatorname{J}( #1 , #2 )}
\newcommandx{\KLLigne}[2]{\operatorname{KL}( #1 | #2 )}
\newcommandx{\KLLignesqrt}[2]{\operatorname{KL}^{1/2}( #1 | #2 )}

\def\gaStep
\def\QKer{Q}

\def\distance{\mathbf{d}}
\newcommandx{\wasserstein}[3][1=\distance,3=]{\mathbf{W}_{#1}^{#3}\left(#2\right)}
\newcommandx{\wassersteinLigne}[3][1=\distance,3=]{\mathbf{W}_{#1}^{#3}(#2)}
\newcommandx{\wassersteinD}[1][1=\distance]{\mathbf{W}_{#1}}
\newcommandx{\wassersteinDLigne}[1][1=\distance]{\mathbf{W}_{#1}}

\def\sigmaD{\sigma^2}

\newcommandx{\phibfs}[1][1=]{\pmb{\varphi}_{\sigmaD_{#1}}}

\newcommandx\sequenceg[3][2=,3=]
{\ifthenelse{\equal{#3}{}}{\ensuremath{( #1_{#2})}}{\ensuremath{( #1_{#2})_{ #2 \geq #3}}}}

\newcommandx{\distV}[1][1=\bfc]{\mathbf{W}_{#1}}
\newcommandx{\distVdeux}[1][1=W_2]{\mathbf{d}_{#1}}

\usepackage[textsize=tiny]{todonotes}

\pgfplotsset{compat=1.16}

\begin{document}

\onecolumn

\icmltitle{Riemannian Diffusion Schr\"odinger Bridge}




\begin{icmlauthorlist}
\icmlauthor{James Thornton}{ox}
\icmlauthor{Michael Hutchinson}{ox}
\icmlauthor{Emile Mathieu}{ox} \\ 
\icmlauthor{Valentin De Bortoli}{ens}
\icmlauthor{Yee Whye Teh}{ox}
\icmlauthor{Arnaud Doucet}{ox}

\end{icmlauthorlist}

\icmlaffiliation{ox}{Department of Statistics,
University of Oxford, UK}
\icmlaffiliation{ens}{Computer Science Department, \\
  ENS, CNRS, PSL University}

\icmlcorrespondingauthor{James Thornton}{james.thornton@stats.ox.ac.uk}

\icmlkeywords{Diffusion processes, Generative modeling, Riemannian manifold, Score-based generative models, Schr\"odinger bridge}

\vskip 0.3in



\printAffiliationsAndNotice{}  % leave blank if no need to mention equal contribution
% Many data modalities are naturally described through a Riemannian geometry.
\begin{abstract}
  Score-based generative models exhibit state of the art performance on density
  estimation and generative modeling tasks.  These models typically assume that
  the data geometry is flat, yet recent extensions have been developed to
  synthesize data living on Riemannian manifolds. Existing methods to accelerate
  sampling of diffusion models are typically not applicable in the Riemannian
  setting and Riemannian score-based methods have not yet been adapted to the
  important task of interpolation of datasets. To overcome these issues, we
  introduce \emph{Riemannian Diffusion Schr\"odinger Bridge}.  Our
  proposed method generalizes Diffusion Schr\"odinger Bridge introduced in
  \cite{debortoli2021neurips} to the non-Euclidean setting and extends
  Riemannian score-based models beyond the first time reversal. We validate our
  proposed method on synthetic data and real Earth and climate data.
%  RDSB accelerates score-based generative modeling on a manifold without closed-form sampling, in addition, RDSB enables dataset interpolation and likelihood computation.

% Score-based generative modeling exhibits state of the art performance for data synthesis and likelihood computation in the Euclidean setting and has recently been generalized to the Riemannian setting. Existing methods to accelerate sampling of diffusion models are typically not applicable in the Riemannian setting, primarily due to the difficulty in sampling diffusions in closed-form on a manifold. To overcome this issue, we  introduce \emph{Riemannian Diffusion Schr\"odinger Bridge} (RDSB). Our proposed method generalizes Diffusion Schr\"odinger Bridge to the non-Euclidean setting and generalizes Riemannian score-based modeling beyond the first time reversal. RDSB accelerates score-based generative modeling on a manifold without closed-form sampling, in addition, RDSB enables dataset interpolation and likelihood computation.  
\end{abstract}

\section{Background}

    \subsection{Score Based Generative Modeling}

    In Euclidean spaces, Score-based Generative Modeling (SGM)
    \citep{song2019generative, song2020score} consists of two main components.
    The first is a forward \emph{noising} process $(\bfX_t)_{t\ge0}$ defined via the
    stochastic differential equation (SDE) \eqref{eq:backward_sde}
    % to samples from data 
    and the initial
    distribution $\bfX_0 \sim \pdata$, targeting an easy-to-sample prior $\pprior$.
    The second component is a backward \emph{denoising} process $(\bfY_t)_{t \geq 0} = (\bfX_{T-t})_{t \in \ccint{0,T}}$ defined by the time-reversal of the noising SDE \eqref{eq:backward_sde}~\citep{cattiaux2021time,haussmann1986time} from $\pprior=p_T$ to $\pdata=p_0$ . Here $f$ is the drift, $g$ is the time-dependent volatility, while $(\bfB_t)_{t\ge0}$ is a d-dimensional Brownian motion and $\Pbb_t$ is the distribution of $\bfX_t$ with corresponding density $p_t$. 
    % $\bfX=\{\bfX_t\}_{t\in[0,T]}$, $\bfX_t \in \mathbb{R}^{d}$.
    The denoising process defines a generative model by sampling
    $\bfY_0 \sim \pprior$
    % \begin{multicols}{2}
    %   \begin{equation}
    %     \rmd \bfX_t = f(t, \bfX_t)  \rmd t + g(t)\rmd\bfB_t%,\quad \bfX_0 \sim p_0
    %  \label{eq:forward_sde} 
    %   \end{equation}
    %   \begin{equation}
    %     \rmd \bfY_t = \{-f(t,\bfY_t) + g^2(t)\nabla \log p_{T-t}(\bfY_t)\}\rmd t + g(t)\rmd \bfB_t \label{eq:backward_sde}
    %   \end{equation}
    % \end{multicols}
    \begin{equation}
     \rmd \bfX_t =  f(t, \bfX_t)  \rmd t + g(t)\rmd\bfB_t , \qquad 
     \rmd \bfY_t =  \{-f(t,\bfY_t) + g^2(t)\nabla \log p_{T-t}(\bfY_t)\}\rmd t + g(t)\rmd \bfB_t.% ,\quad \bfY_0 \sim p_T  && \hfill
     \label{eq:backward_sde}
    \end{equation}
    The intractable score term, $\nabla \log p_{T-t}(\bfY_t)$, may be
    approximated by using the following score matching identity
    $\textstyle{\nabla_{x_t} \log p_t(x_t) = \int_{\rset^d} \nabla_{x_t} \log
      p_{t|0}(x_t|x_0)~p_{0|t}(x_0|x_t) \rmd x_0}$, with tractable transition
    density $p_{t|0}$.  This is then used to train a neural network
    $\bm{s}_\theta: \mathbb{R} \times \mathbb{R}^{d} \rightarrow \mathbb{R}^{d}$
    with regression objective
    $\bm{s}_{\theta^*}= \arg\min_\theta \mathbb{E}_{\bfX_t,\bfX_0}
    \normLigne{\nabla_x \log p_{t|0}(\bfX_t|\bfX_0) - \bm{s}_\theta(t,
      \bfX_t)}^2$.  Once trained, one can generate samples which are
    approximately distributed according to $\pdata$ via \eqref{eq:backward_sde}
    (e.g.\ considering the Euler--Maruyama discretization of this SDE) by 
    substituting $\bm{s}_{\theta^*}(t, x_t) \approx \nabla_{x_t} \log
    p_t(x_t)$. % and $\bfY_0\sim \pprior$.

    \subsection{Riemannian Score Based Generative Modeling}

    % \begin{wrapfigure}{R}{0.5\linewidth}
    % \begin{align}
    %   \rmd \bfX_t &= f(t,\bfX_t) \rmd t + g(t)\rmd \bfB_t^\M
    %  \label{eq:rforward_sde} \\
    %  \rmd \bfY_t &= \{-f(t,\bfY_t) + \nabla \log p_{T-t}(\bfY_t)\} \rmd t + g(t)\rmd \tilde{\bfB}_t^\M
    %  \label{eq:rbackward_sde}
    % \end{align}
    % \end{wrapfigure}

    \citet{de2022riemannian} extended SGM to compact Riemannian manifolds,
    denoted $\M$, henceforth abbreviated RSGM for Riemannian Score-based
    Generative Modeling.  Given a $\M$-valued diffusion process (\ref{eq:rbackward_sde}, left), where $\bfB_t^\M$ denotes Brownian motion on $\M$; the time
    reversal process may be written as (\ref{eq:rbackward_sde}, right)~\citep[Theorem
    1]{de2022riemannian}
    \vspace{-0.1cm}
    \begin{align}
      \rmd \bfX_t &= f(t,\bfX_t) \rmd t + g(t)\rmd \bfB_t^\M,
 &
    \rmd \bfY_t &= \{-f(t,\bfY_t) + g^2(t)\nabla \log p_{T-t}(\bfY_t)\} \rmd t + g(t)\rmd \tilde{\bfB}_t^\M.
    \vspace{-0.1cm}
     \label{eq:rbackward_sde}
    \end{align}
     For a more thorough background on Riemannian geometry
    and time-reversal on manifolds see \citet[App. B and G]{de2022riemannian}.
     Simulating a diffusion on a manifold is crucial to this approach.
     In contrast to Euclidean SGMs, closed-form sampling schemes are generally not available for diffusions on manifolds, hence approximation schemes such as the Geodesic Random Walk from Algo. \ref{alg:grw} are required. Note that $\bfB_t^\M$ may be simulated using Algo \ref{alg:grw} for $g=1,f=0$.
    %  and unlike Euclidean SGM, may only be performed approximately using \Cref{alg:grw}, .
     This consists of applying an Euler--Maruyama step in the tangent space,
     whereby Gaussian noise has also been projected to the tangent space using
     the projection operator $\mathrm{P}$, then projecting back to the manifold
     using the exponential mapping.
    %  Unlike for SGM, it is not possible, in general, to sample $\bfX_t|\bfX_0$ in closed form and one must instead simulate a diffusion step-wise.

\begin{figure}[ht]
        \centering
\begin{tikzpicture}
        % forward 1 ------------------------------------------------------------------
        \node[inner sep=0pt, label={\small }] (f1_T) at (\diaW,0)
           {\includegraphics[trim=30 40 40 40, clip,width=\imW\textwidth]{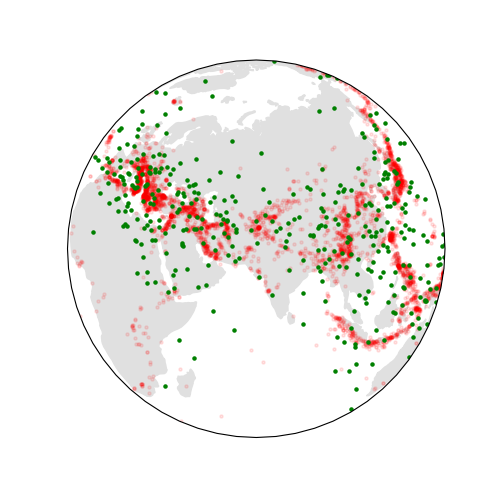}};
        \node[inner sep=0pt, label={\small },opacity=\imOp] (f1_1) at (\diaW/4,0)
           {\includegraphics[trim=30 40 40 40, clip,width=\imW\textwidth]{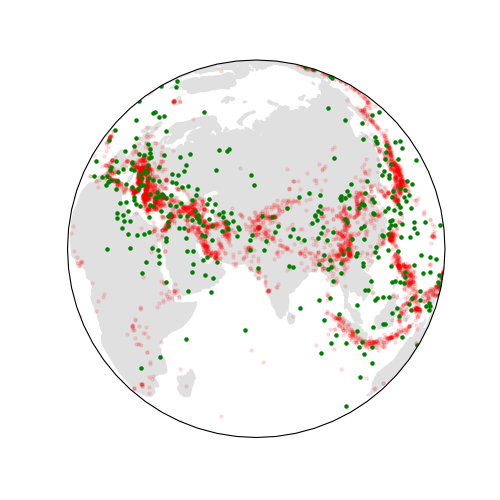}};
        \node[inner sep=0pt, label={\small },opacity=\imOp] (f1_2) at (\diaW/2,0)
           {\includegraphics[trim=30 40 40 40, clip,width=\imW\textwidth]{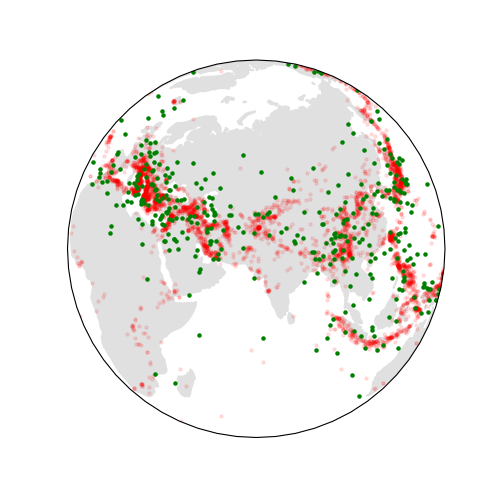}};
        \node[inner sep=0pt, label={\small },opacity=\imOp] (f1_3) at (3*\diaW/4,0)
            {\includegraphics[trim=30 40 40 40, clip,width=\imW\textwidth]{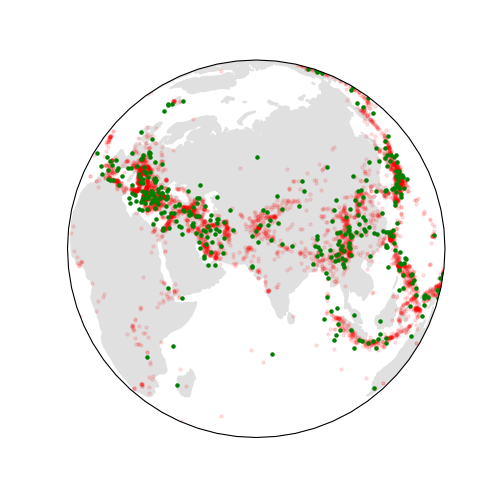}};
        \node[inner sep=0pt, label={\small }] (f1_data) at (0,0)
            {\includegraphics[trim=30 40 40 40, clip,width=\imW\textwidth]{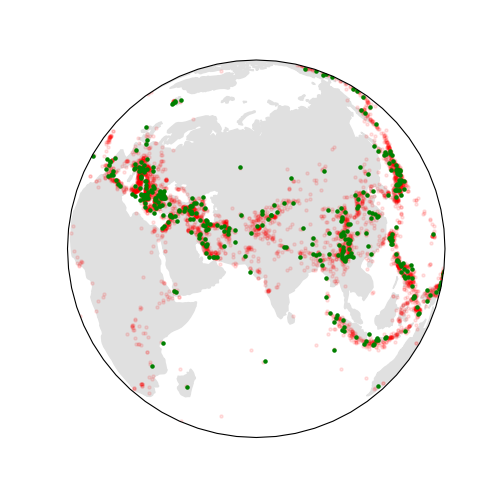}};
        
        % backward 1 ------------------------------------------------------------------    
        \node[inner sep=0pt, label={\small }] (b1_data) at (0,-\nextRow)    
            {\includegraphics[trim=30 40 40 40, clip,width=\imW\textwidth]{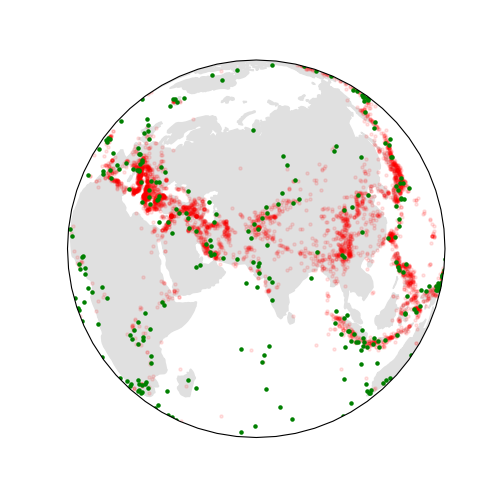}};
        \node[inner sep=0pt, label={\small },opacity=\imOp] (b1_1) at (\diaW/4,-\nextRow)
            {\includegraphics[trim=30 40 40 40, clip,width=\imW\textwidth]{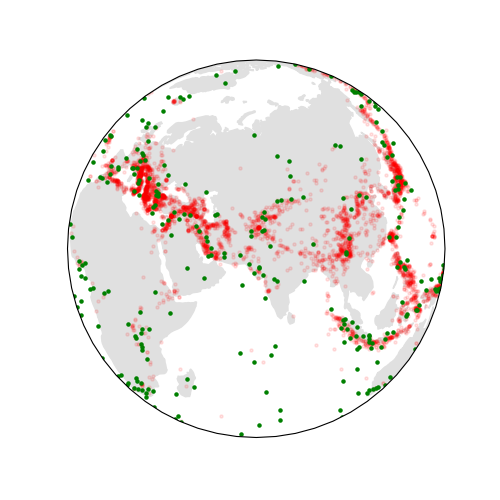}};
        \node[inner sep=0pt, label={\small },opacity=\imOp] (b1_2) at (2*\diaW/4,-\nextRow)
            {\includegraphics[trim=30 40 40 40, clip,width=\imW\textwidth]{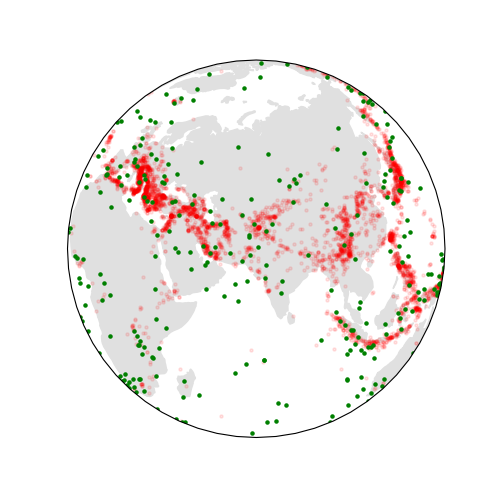}};
        \node[inner sep=0pt, label={\small },opacity=\imOp] (b1_3) at (3*\diaW/4,-\nextRow)
            {\includegraphics[trim=30 40 40 40, clip,width=\imW\textwidth]{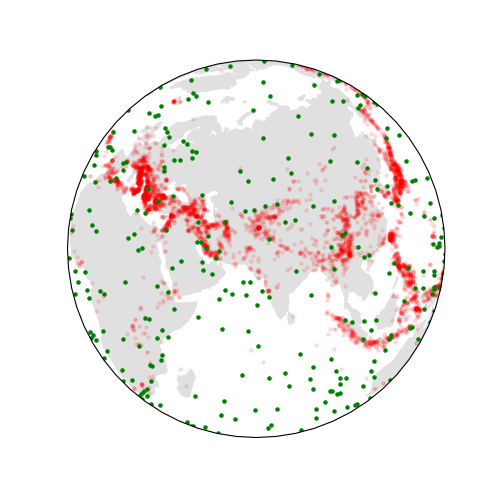}};
        \node[inner sep=0pt, label={\small }] (b1_T) at (\diaW,-\nextRow)
            {\includegraphics[trim=30 40 40 40, clip,width=\imW\textwidth]{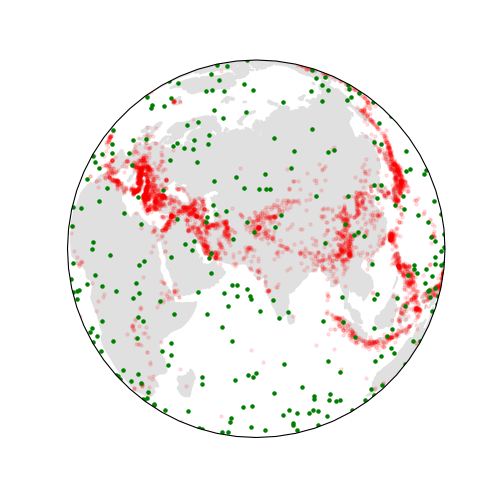}};
        
        % forward 15 ------------------------------------------------------------------
        \node[inner sep=0pt, label={\small }] (f15_data) at (0,-\diaJump)
            {\includegraphics[trim=30 40 40 40, clip,width=\imW\textwidth]{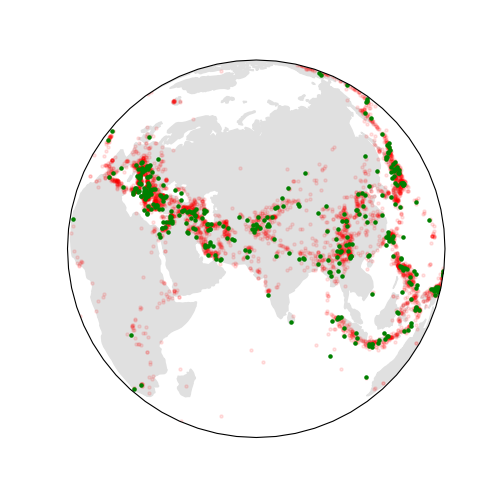}};
        \node[inner sep=0pt, label={\small },opacity=\imOp] (f15_1) at (\diaW/4,-\diaJump)
            {\includegraphics[trim=30 40 40 40, clip,width=\imW\textwidth]{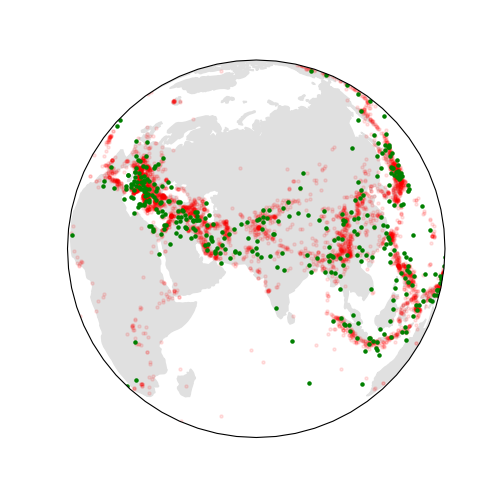}};
        \node[inner sep=0pt, label={\small },opacity=\imOp] (f15_2) at (\diaW/2,-\diaJump)
           {\includegraphics[trim=30 40 40 40, clip,width=\imW\textwidth]{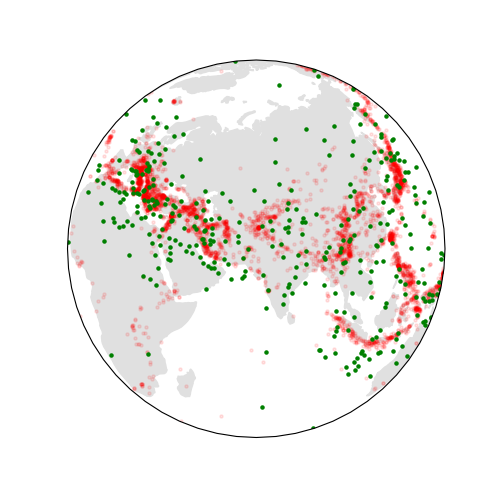}};
        \node[inner sep=0pt, label={\small },opacity=\imOp] (f15_3) at (3*\diaW/4,-\diaJump)
            {\includegraphics[trim=30 40 40 40, clip,width=\imW\textwidth]{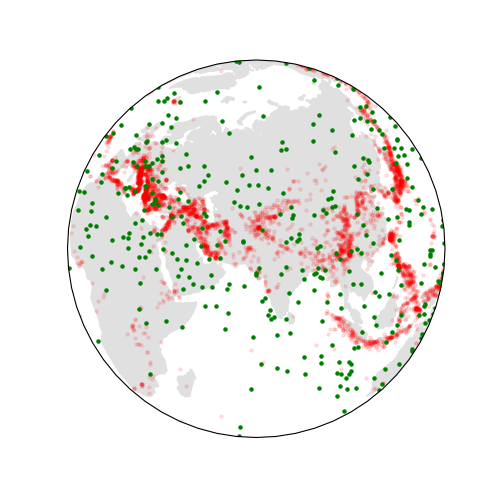}};
        \node[inner sep=0pt, label={\small }] (f15_T) at (\diaW,-\diaJump)
            {\includegraphics[trim=30 40 40 40, clip,width=\imW\textwidth]{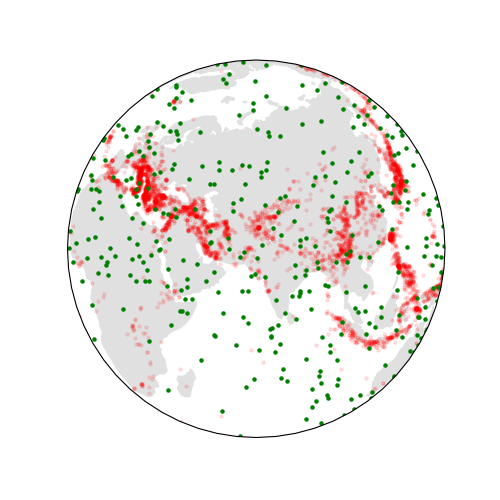}};
        % backward 15 ------------------------------------------------------------------
        \node[inner sep=0pt, label={\small }] (b15_data) at (0,-\diaJump-\nextRow)
           {\includegraphics[trim=30 40 40 40, clip,width=\imW\textwidth]{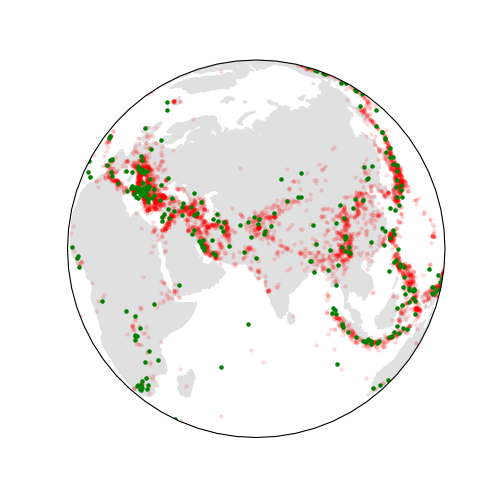}};
        \node[inner sep=0pt, label={\small },opacity=\imOp] (b15_1) at (\diaW/4,-\diaJump-\nextRow)
           {\includegraphics[trim=30 40 40 40, clip,width=\imW\textwidth]{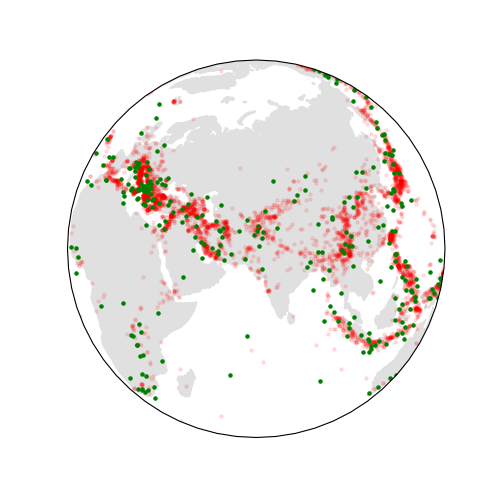}};
        \node[inner sep=0pt, label={\small }, opacity=\imOp] (b15_2) at (\diaW/2,-\diaJump-\nextRow)
           {\includegraphics[trim=30 40 40 40, clip,width=\imW\textwidth]{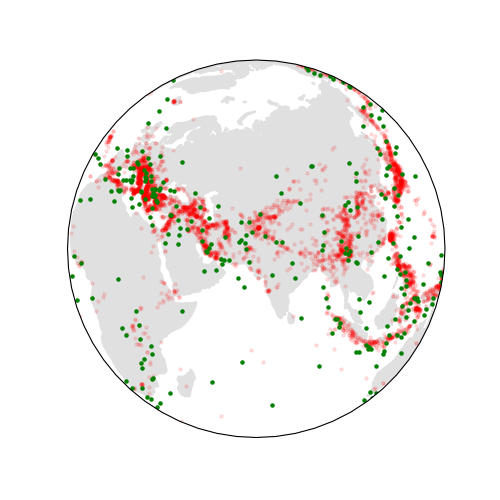}};
        
        \node[inner sep=0pt, label={\small }, opacity=\imOp] (b15_3) at (3*\diaW/4,-\diaJump-\nextRow)
            {\includegraphics[trim=30 40 40 40, clip,width=\imW\textwidth]{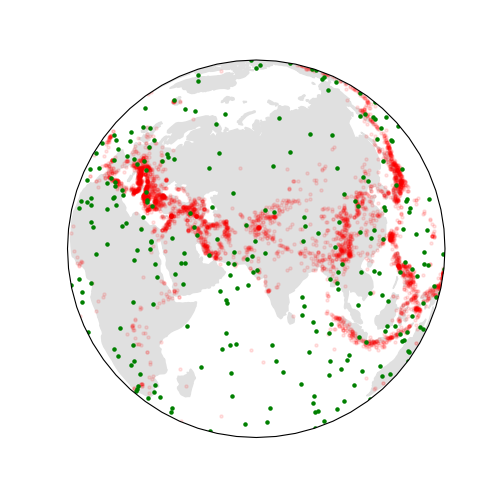}};
            
        \node[inner sep=0pt, label={\small }] (b15_T) at (\diaW,-\diaJump-\nextRow)
           {\includegraphics[trim=30 40 40 40, clip,width=\imW\textwidth]{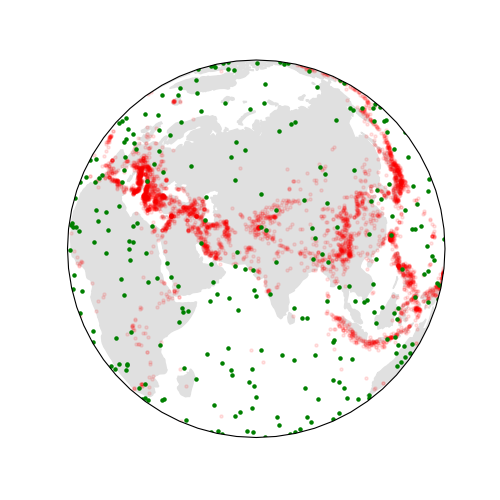}};
        
        % arrows 15 ------------------------------------------------------------------
        % \draw[-{Latex[length=2mm]},dashed] (\diaW/2,-1.25*\diaH/4) -- (\diaW/2,-\diaJump+0.5) ;
        % \draw[-{Latex[length=2mm]},dashed] (\diaW/4,-1.25*\diaH/4) -- (\diaW/4,-\diaJump+0.5) ;
        % \draw[-{Latex[length=2mm]},dashed] (3*\diaW/4,-1.25*\diaH/4) -- (3*\diaW/4,-\diaJump+0.5) ;

        \path[-{Latex[length=3mm]}, thick]
         (f15_data) edge[bend left=\bend] node [fill=white,xshift=-40pt] {Forward} (f15_T)
         (f1_data) edge[bend left=\bend] node [fill=white,xshift=-40pt] {Forward} (f1_T)
         (b1_T) edge[bend left=\bend] node [fill=white,xshift=-40pt] {Backward} (b1_data)
         (b15_T) edge[bend left=\bend] node [fill=white,xshift=-40pt] {Backward} (b15_data);

        \draw[-{Latex[length=3mm]}, very thick] (-1,-\diaH) -- (\diaW,-\diaH)
            node[midway,fill=white] {$t$}
            node[pos=0.1,below,fill=white] {$t=0$}
            node[pos=0.9,below,fill=white] {$t=T$};
        
        % \draw[red,thick] ($(f15_data.north west)+(-0.05,0.05)$)  rectangle ($(b15_T.south east)+(0.05,-0.05)$)
        % node[pos=0.6, left,fill=white] {Interpolation Model};
        
        \draw[-{Latex[length=3mm]},very thick] (-1.2,0) -- (-1.2,-\diaH)
            % node[pos=0.3, left, fill=white] {DSB \\ Steps}
            node[pos=0.1, fill=white] {RSGM}
            node[pos=0.7,fill=white] {RDSB};
      \end{tikzpicture}
              \caption{Earthquake data: empirical data in red, generated samples in green using $N=10$ diffusion steps. Top forward/backward pair: RSGM. Bottom pair: RDSB, with $5$ IPF steps.}
       \label{fig:schro_bridge}          
     \end{figure}
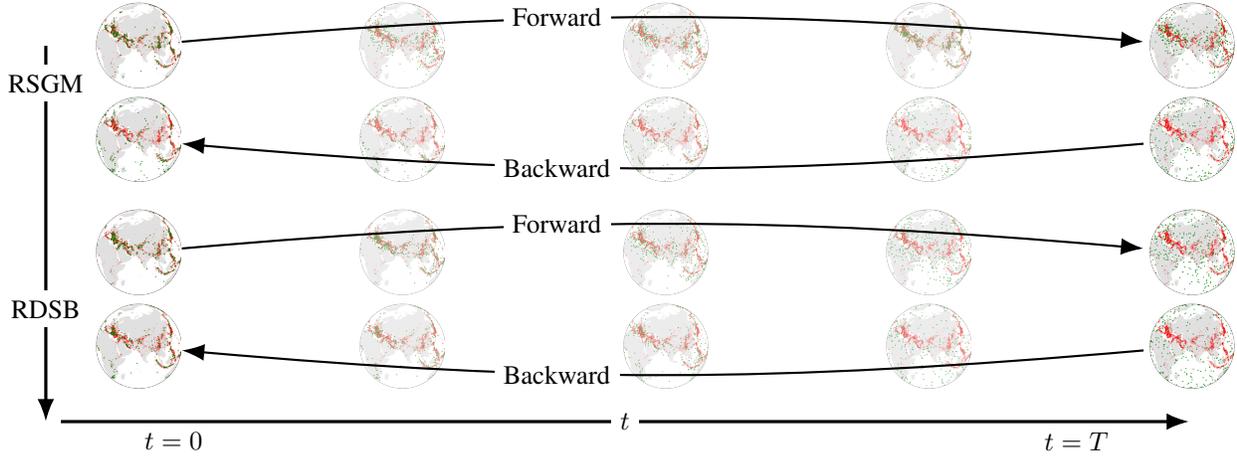
     \begin{wrapfigure}{r}{.4\linewidth}
    \vspace{-.5cm}
    % \begin{align}
    %   \rmd \bfX_t &= f(t,\bfX_t) \rmd t + g(t)\rmd \bfB_t^\M
    %  \label{eq:rforward_sde} \\
    %  \rmd \bfY_t &= \{-f(t,\bfY_t) + g^2(t)\nabla \log p_{T-t}(\bfY_t)\} \rmd t + g(t)\rmd \tilde{\bfB}_t^\M
    %  \label{eq:rbackward_sde}
    % \end{align}
    \begin{minipage}{.4\textwidth}
    \centering
    \begin{algorithm}[H]
    \caption{\small Simulating diffusions on a manifold}
    \label{alg:grw}
    \begin{algorithmic}[1]
     \small
      \STATE \textbf{Input:} step size: $\gamma$ , initial state $X_0$ 
      \FOR{$k \in \{0, \dots, N-1\}$}
      \STATE $\bar{\bfZ}_{k+1} \sim \mathcal{N}(0, I_p), \quad \bfZ_{k+1} = \mathrm{P}(\bfX_k) \bar{\bfZ}_{k+1}$ 
      \STATE $\bfW_{k+1} = \gamma f(k \gamma, \bfX_k) + \sqrt{\gamma} g(k \gamma) \bfZ_{k+1}$      \STATE $\bfX_{k+1} = \exp_{\bfX_k}\left(\bfW_{k+1}\right)$ 
      \ENDFOR
      \STATE {\bfseries return} $\{\bfX_k\}_{k=0}^{N-1}$
    \end{algorithmic}
    \end{algorithm}
    \end{minipage}
    \vspace{-.5cm}
    \end{wrapfigure}
    Score matching may be used to approximate the time-reversal
    \eqref{eq:rbackward_sde} in the Riemannian setting
    \citep{de2022riemannian}. Unlike for Euclidean SGM,
    $\nabla \log p_{t|0}(x_t|x_0)$ is not available in closed-form. Instead, one
    may use the score identity \footnote{Here all gradients are considered w.r.t. the  Riemannian metric of $\M$.}
    $\textstyle{ \nabla_{x_t} \log p_t(x_t) = \int_{\M} \nabla_{x_t} \log
      p_{t|s}(x_t|x_s) \Pbb_{s|t}(x_t, \rmd x_s) }$ for $s\approx t$, where
    again $\Pbb$ is the distribution of $\bfX$ and $p_t$ corresponds to the density of $\Pbb_t$ 
    with respect to the uniform distribution on $\M$, then
    $\nabla \log p_t(x_t) = \argmin_{\mathbf{s}_t} \int_{\M^2}
    \normLigne{\nabla_x \log p_{t|s}(x_t|x_s) - \mathbf{s}_t(x_t)}^2 \rmd
    \Pbb_{s,t}(x_s,x_t)$.

    \subsection{The Schr\"odinger Bridge Problem} \label{sec:sbp}

    A Schr\"odinger bridge extension of SGM  has been introduced to reduce the number of diffusion steps for SGM by learning \emph{both} forward and backward diffusions \citep{debortoli2021neurips}.
    We briefly recall the notion of dynamical Schr\"odinger bridge \citep{leonard2012schrodinger,chen2016entropic,vargas2021solving,debortoli2021neurips,chen2021likelihood}. We consider a reference path probability measure 
    $\Pbb \in \Pens(\rmc(\ccint{0,T}), \M)$ where $\Pens(\rmc(\ccint{0,T}), \M)$ is the space measures on continuous paths $(\bfX_t)_{t\in[0,1]}$ in $\M$ . In practice, we set $\Pbb$ to be the
    distribution of the Brownian motion $(\bfB_t^\M)_{t \in \ccint{0,T}}$ initialized from $\pdata$, i.e.
    $\bfB_0^\M$ has distribution $\pdata$. We consider
    the \emph{dynamical Schr\"odinger bridge problem}
    \begin{equation}
      \Qbb^\star = \argmin \ensembleLigne{\KL{\Qbb}{\Pbb}}{\Qbb \in \Pens(\rmc(\ccint{0,T}), \M), \ \Qbb_0 = \pdata, \ \Qbb_T = \pprior} . 
    \end{equation}
    The idealised solution $\Qbb^\star$ is called the Schr\"odinger Bridge (SB).
    Given a backward process
    $(\bfY_t^\star)_{t \in \ccint{0,T}}$ associated to
    $\Qbb^\star$, one can obtain a generative model as
    follows. First sample from $\bfY^\star_T \sim \pprior$ and then follow
    the (backward) dynamics of $(\bfY^\star_t)_{t \in \ccint{0,T}}$. By definition, we obtain that $\bfY^\star_0 \sim \pdata$, the data distribution.
    
    In practice, the solution of the SB problem is approximated using the Iterative Proportional Fitting (IPF) algorithm, which coincides with the Sinhkorn algorithm in discrete space \citep{sinkhorn1967diagonal,peyre2019computational}. IPF defines a sequence of path probability measures $(\Qbb^n)_{n \in \nset} \in (\Pens(\rmc(\ccint{0,T}, \M)))^\nset$, such that $\Qbb^0 = \Pbb$ and for any $n \in \nset$
    \begin{align}
      &\Qbb^{2n+1} = \argmin \ensembleLigne{\KL{\Qbb}{\Qbb^{2n}}}{\Qbb \in \Pens(\rmc(\ccint{0,T}), \M), \Qbb_T = \pprior}  , \\
      &\Qbb^{2n+2} = \argmin \ensembleLigne{\KL{\Qbb}{\Qbb^{2n+1}}}{\Qbb \in \Pens(\rmc(\ccint{0,T}), \M), \Qbb_0 = \pdata} .
    \end{align}
    Under mild assumptions on $\Pbb$, $\pdata$ and $\pprior$, we have that
    $(\Qbb^n)_{n \in \nset}$ converges towards $\Qbb^\star$ \cite{nutz2022stability}.

\section{Riemannian Diffusion Schr\"odinger Bridge}

     In Euclidean state spaces, \citet{debortoli2021neurips} proposed  Diffusion Schr\"odinger Bridge (DSB), an algorithm to approximate the solution to the SB problem based on time-reversal, using score matching to approximate the IPF iterates. We propose Riemannian Diffusion Schr\"odinger Bridge (RDSB), an extension of DSB to approximate solutions of the SB problem for compact Riemannian manifolds. 
%    \subsection{Iterative Proportional Fitting on a Compact Manifold}
    
    %     \begin{wrapfigure}{R}{.35\linewidth}
    %     \vspace{-0.5cm}
    %     \begin{align}
    %   \rmd \bfX^n_t &= f^n_t(\bfX_t) \rmd t + g(t)\rmd \bfB_t^\M
    %  \label{eq:drift_forward_sde} \\
    %  \rmd \bfY^n_t &= b^n_t(\bfY_t) \rmd t + g(t)\rmd \tilde{\bfB}_t^\M
    %  \label{eq:drift_backward_sde}
    % \end{align}
    % \end{wrapfigure}
    We first connect the IPF iterates
    $(\Qbb^n)_{n \in \nset}$ with time reversal of diffusion processes on $\M$. To simplify notation, we rewrite \cref{eq:rbackward_sde} in terms of drift functions $f^n$ and $b^n$ in \cref{eq:drift_forward_sde,eq:drift_backward_sde}. Here $\bfY^{n}$ corresponds to the time-reversal of $\bfX^{n}$, initialized at $\bfY^{n}_T \sim \pprior$.  $\bfX^0$ is the Brownian motion on $\M$, $f^n=0$, and  $\bfX^{n+1}$ denotes the time-reversal of the diffusion $\bfY^{n}$, initialized at $\bfX^{n}_0 \sim \pdata$.
    \vspace{-1cm}
    \begin{multicols}{2}
      \begin{equation}
        \rmd \bfX^n_t = f^n(t,\bfX_t^n) \rmd t + g(t)\rmd \bfB_t^\M , 
         \label{eq:drift_forward_sde}
      \end{equation}\break
      \begin{equation}
        \rmd \bfY^n_t = b^n(T-t,\bfY_t^n) \rmd t + g(t)\rmd \tilde{\bfB}_t^\M .
         \label{eq:drift_backward_sde}
      \end{equation}
    \end{multicols}

    \begin{wrapfigure}{R}{.42\linewidth}
    \vspace{-.4cm}
    \begin{minipage}{0.42\textwidth}
    \begin{algorithm}[H]
        \caption{RDSB}
        \label{algo:ipf_score}
        \begin{algorithmic}[1] 
          \FOR{$n \in \{0, \dots,L\}$}
          \WHILE{not converged}
          \STATE Sample $t_i \sim \textrm{Uniform}([0,T])$
          \STATE Simulate $\{\bfX^i_{t_i}\}_{i=0}^{B}$, where  $\bfX^i_0 \sim \pdata$
            \STATE Compute $\hlb_n(\phi^n)$ using \eqref{eq:backward_loss}
            \STATE $\phi^{n} \leftarrow \textrm{Gradient Step}(\hlb_n(\phi^n))$ 
          \ENDWHILE 
          \WHILE{not converged}
            \STATE Sample $t_i \sim \textrm{Uniform}([0,T])$
            \STATE Simulate $\{\bfY^i_{t_i}\}_{i=0}^{B}$, where $\bfY^i_T \sim \pprior$
          \STATE Compute $\hlf_{n+1}(\theta^{n+1})$ using \eqref{eq:forward_loss}
          \STATE $\theta^{n+1} \leftarrow \textrm{Gradient Step}(\hlf_{n+1}(\theta^{n+1}))$
          \ENDWHILE 
          \ENDFOR 
          \STATE 
          \textbf{Output: } $(\theta^{L+1}, \phi^{L})$
        \end{algorithmic}
      \end{algorithm}
      \end{minipage}
      \vspace{-.8cm}
    \end{wrapfigure}
    \begin{proposition}
      \label{prop:continuous_schro}
      Let $\Pbb$ be the path measure of the Brownian motion initialized at
      $\pprior$ and IPF iterates $(\Qbb^n)_n$ be as defined in \Cref{sec:sbp}.
      Assume that for any $n \in \nset$, $\KL{\Qbb^n}{\Pbb}< +\infty$ and that
      for any $t \in \ccint{0,T}$ and $n \in \nset$, $\Qbb^n_t$ admits a smooth
      positive density w.r.t.\ $\pprior$. Then, for any $n \in \nset$:
      $\Qbb^{2n}$ and $R(\Qbb^{2n+1})$ solve the time-reversal for
      \eqref{eq:drift_forward_sde} and \eqref{eq:drift_backward_sde}
      respectively.  $R(\Pbb)_t=\Pbb_{T-t}$ denotes the reverse time and for any
      $n \in \nset$, $t \in \ccint{0,T}$ and $x \in \M$,
      $b^{n}(t,x) = -f^{n}(t, x) + g(t)^2\nabla \log p^{n}_t(x)$,
      $f^{n+1}(t,x) = -b^n(t,x) + g(t)^2\nabla \log q^n_t(x)$, with
      $f^0(t,x) = 0$, and $p^n_t$, $q_t^n$ the densities of $\Qbb^{2n}_t$ and
      $\Qbb_t^{2n+1}$.
    \end{proposition}
    \vspace{-0.2cm}
    \textit{Proof}
      The proof follows \citet[Proposition 6]{debortoli2021neurips} using
      \citet[Theorem 1]{de2022riemannian} instead of \citet[Theorem
      4.19]{cattiaux2021time}

    In particular, we have that $\Qbb^1$ is the diffusion process associated with
    RSGM, \ie \ the time-reversal of the Brownian motion initialized at
    $\pprior$. Hence, $\Qbb^{2n+1}$ for $n \in \nset$ with $n \geq 1$ can be seen as
    a refinement of $\Qbb^1$. In the next proposition, we show that the drift term
    of the diffusion processes associated with $(\Qbb^n)_{n \in \nset}$ can be
    approximated leveraging score-based techniques.
    
    \begin{proposition}
      \label{prop:loss_implicit_explicit}
      Let $(\bfX_t)_{t \in \ccint{0,T}}$ be a $\M$-valued process with
      distribution $\Pbb \in \Pens(\rmc(\ccint{0,T}), \M)$ such that for any
      $t \in \ccint{0,T}$, $\bfX_t$ admits a positive density
      $p_t \in \rmc^\infty(\M)$ w.r.t.\ $\pprior$.  For any $t \in \ccint{0,T}$
      and $x \in \M$, let $b(t,x) = -f(t,x) + g(t)^2\nabla \log p_t(x)$.  Then,
      for any $t \in \ccint{0,T}$, we have that \vspace{-0.2cm}
      \[
        b(t, \cdot) = \argmin_{r \in \mathrm{L}^2(\Pbb_t)} \expeLigne{\tfrac{1}{2}\normLigne{f(t, \bfX_t) + r(\bfX_t)}_2^2 + g(t)^2\dive(r)(\bfX_t)}. \notag 
      \]
    \end{proposition}
    \vspace{-0.3pt}
    \textit{Proof.} For any $x \in \M$, $t \in \ccint{0,T}$,
      $b(t, \cdot) = \argmin_{r\in \XM}\expeLigne{\normLigne{r(t,\bfX_t) - \{-f(t,\bfX_t) + g(t)^2\nabla \log p_t(\bfX_t) \}}}_2^2 $. 
        Expanding the quadratic, dropping terms not dependent on $r$ and using the divergence theorem,  \citep[see][p.51]{lee2018introduction}, $\expeLigne{\langle r(\bfX_t), \nabla \log p_t(\bfX_t) \rangle_\M} = -\expeLigne{\dive(r)(\bfX_t)}$ concludes the proof. 

    \begin{wrapfigure}{r}{0.6\linewidth}
    \vspace{-1cm}
    \begin{align}
        \hat{\ell}^b_n(\phi)&\textstyle{=\sum_{i=1}^B\tfrac{1}{2}\normLigne{f^{n}_\theta(t_{i}, x^i_{t_i}) + b^n_\phi(t_{i},x^i_{t_i}))}_2^2 + g^2(t_{i})\dive(b^n_\phi)(t_{i}, x^i_{t_i})}
         \label{eq:backward_loss}\\
        \hat{\ell}^f_n(\theta)&\textstyle{=\sum_{i=1}^B\tfrac{1}{2}\normLigne{b^{n-1}_\phi(t_{i}, x^i_{t_i}) + f^n_\theta(t_{i},x^i_{t_i})}_2^2 + g^2(t_{i})\dive(f^n_\theta)(t_{i}, x^i_{t_i})}
         \label{eq:forward_loss}
    \end{align}
    \vspace{-.8cm}
        \end{wrapfigure}
        
        In practice, for $L$ IPF steps, $(f^n)_{n=1}^L$ and $(b^n)_{n=0}^L$ are approximated using neural networks,  $(f^n_{\theta^n}   )_{n=1}^L$, $(b^n_{\phi^n})_{n=1}^L$ with parameters $(\theta^n, \phi^n)_n$. Approximating drifts or means of the SDE is computationally cheaper than storing an evaluating $2L$ separate score networks. \Cref{prop:loss_implicit_explicit} provides loss functions. The divergence terms may be estimated using automatic differentiation or Hutchinson's trace estimator \citep{hutchinson1989stochastic}.
        %
    %     \begin{figure}[ht]
    %         \centering
    %         \includegraphics[trim=30 50 50 50, clip, width=0.24\linewidth]{plots/convergence/b0_final.png}
    %         \includegraphics[trim=30 50 50 50, clip, width=0.24\linewidth]{plots/convergence/b2_final.png}
    %         \includegraphics[trim=30 50 50 50, clip, width=0.24\linewidth]{plots/convergence/b4_final.png}
    %     \caption{Generated samples across observed after multiple IPF iterations.}
    %     \label{fig:dsb_convergence}
    % \end{figure}
        %         % \includegraphics[trim=30 30 30 30, clip, width=0.15\linewidth]{plots/earthquake_interp/earthquake_backward_0_8.png}
    %         % \hspace{0.2cm}
    %         % \includegraphics[trim=30 30 30 30, clip, width=0.15\linewidth]{plots/earth/fire_backward_0_8.png}
    %         % \hspace{0.2cm}
    %         % \includegraphics[trim=30 30 30 30, clip, width=0.15\linewidth]{plots/earth/vol_backward_0_8.png}
    %         % \hspace{0.2cm}
    %         %  \includegraphics[trim=30 30 30 30, clip, width=0.15\linewidth]{plots/earth/flood_backward_0_8.png}
    %         % \\
    \begin{figure}[ht]
            \centering
            \subfigure[Earthquake]{\includegraphics[trim=30 30 30 30, clip, width=0.22\linewidth]{plots/earthquake_interp/earthquake_backward_5_9.png}}
             \subfigure[Fire]{\includegraphics[trim=30 30 30 30, clip, width=0.22\linewidth]{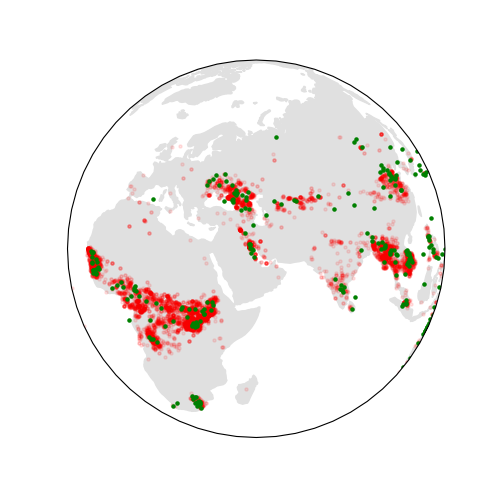}   }         
             \subfigure[Volcano]{\includegraphics[trim=30 30 30 30, clip, width=0.22\linewidth]{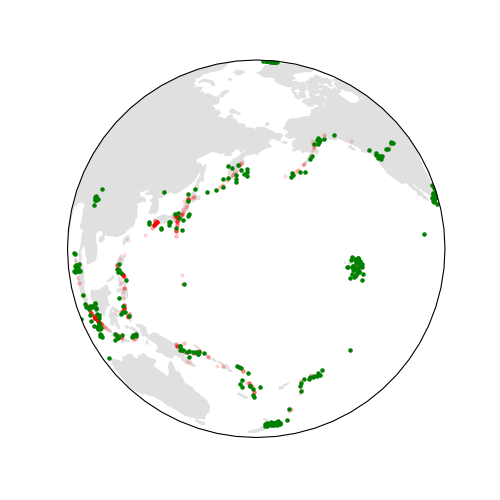}}
             \subfigure[Flood]{\includegraphics[trim=30 30 30 30, clip, width=0.22\linewidth]{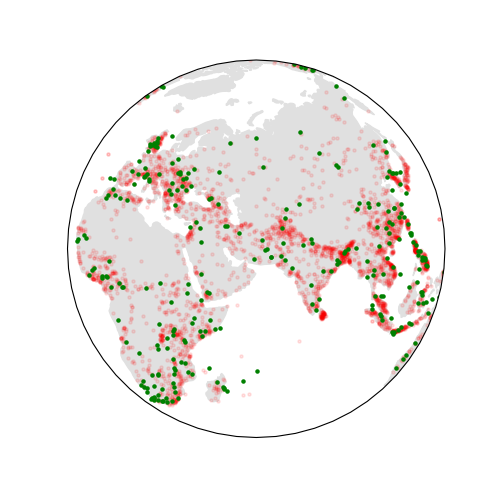}}
             \caption{Climate data: $N=10$ steps, $5$ IPF iterations. Red points
               are training data and green are generated points.}
        \label{fig:earth}
    \end{figure}
    
    \textbf{Likelihood Computation.} The Ordinary Differential Equation (ODE),
    $\rmd \bfhX_t = \{f(t, \bfhX_t) - \tfrac{1}{2} g(t)^2\nabla \log
    p_{T-t}(\bfhX_t)\} \rmd t$, has the same marginal probabilities as SDE
    \eqref{eq:backward_sde}, in particular $\pdata$ and hence may be used for
    likelihood computation using an adaptive ODE solver and trained score
    \citep{song2020score}.  Similar results hold for the Schr\"odinger Bridge
    \citep{debortoli2021neurips, chen2021likelihood} and in the Riemannian
    setting \citep{de2022riemannian}. We construct the appropriate probability
    flow ODE using drift approximations as
    $\rmd \bfhX_t = \frac{1}{2}\{f_\theta(t,\bfhX_t)-b_\phi(t,\bfhX_t)\} \rmd
    t$. The computed likelihood assumes convergence of the forward noising
    process to be valid, in particular that $p_T = \pprior$. RDSB enforces this
    convergence.
    
    % \textbf{Corrector step.}
    % The corrector framework of \citet{song2020score} may also be used within RDSB, following the Riemannian extension of \citet{de2022riemannian}. The corrector step entails simulating an intermediate SDE within invariant measure $p_{T-t}$ at sampling time. For each $t$, this intermediate SDE is $\bfhX^s_t =  \tfrac{1}{2} \nabla \log p_{T-t}(\bfhX_s) \rmd s + \rmd \bfB_s^\M$. Similar to above, this may be approximated as $\bfhX_s =  \tfrac{1}{2g(t)^2} (f_\theta(t,\bfhX_t)+b_\phi(t,\bfhX_t)) \rmd s + \rmd \bfB_s^\M$.
    
    \textbf{Other acceleration methods.}
    Leading SGM acceleration techniques \citep{xiao2021tackling, song2020denoising} use an implicit approach to denoising by estimating $x_0$ from $x_t$, then sampling $\bfX_{s}|x_t, x_0$ for $s<t$. These techniques are not applicable in the Riemannian setting as it is not typically possible to sample $\bfX_s|x_t, x_0$ for $t>s$ or $\bfX_0|x_t$ for large jumps $t\not\approx s$, $t\not\approx 0$.

    \section{Experiments}
    In each of the experiments, $4$-layer, fully connected networks with hidden
    width $512$ are used for both $f_\theta$, $b_\phi$. $N=10$ diffusion steps
    of size $1/N$ was used. Triangular schedule $g^2(t)$ was selected, linearly
    interpolated from a peak of $0.05$ at $t=T/2$ and low of $0.001$ at
    $t=0,T$. Other experimental details follow \citet{de2022riemannian}.
    
    \subsection{Earth and Climate Data}
        \begin{minipage}[t]{0.45\linewidth}
      We validate RDSB on empirical distributions of occurrences of real Earth and climate science events including earthquakes \citep{earthquake_dataset}; wild
    fires \citep{fire_dataset}; volcanic eruptions \citep{volcanoe_dataset}, and floods \citep{flood_dataset}. \Cref{fig:earth} illustrates that generated samples are visually similar to the empirical datasets. 
    
    Close inspection of \Cref{fig:schro_bridge} shows RDSB with $5$ IPF iterations exhibits better convergence than RSGM (equivalent to RDSB with $1$ IPF iteration) for $N=10$ diffusion steps on the earthquake data. In particular, RDSB generates fewer outlying samples to true data samples.
     \end{minipage}
     \begin{minipage}[t]{0.5\linewidth}
     \vspace{-1.4cm}
     \begin{figure}[H]
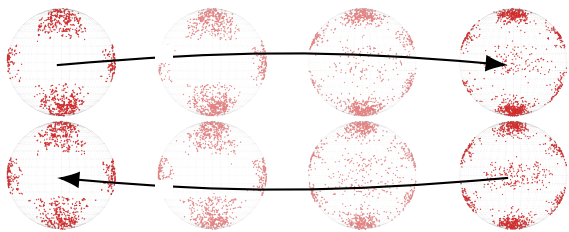

     \include{plots/spherical_diagram}
     \vspace{-0.5cm}
      \caption{Interpolation between spherical harmonic datasets $(l=2,m=4)$ and $(l=2,m=4)$ after $2$ IPF iterations}
      \label{fig:spherical_interp}          
     \end{figure}
     \end{minipage}
    
    \subsection{Dataset Interpolation on a Manifold}
    RDSB enables interpolation between datasets by choosing $\pprior$ to be
    another dataset, and not necessary $p_T$, the terminal distribution of the
    inital noising process. \Cref{fig:spherical_interp} illustrates RDSB applied
    to samples on the sphere taken with probability proportional to the real
    component of spherical harmonic function with parameters $(l=2,m=4)$ and
    $(l=2,m=6)$.

    % dataset interpolation
    
    % \begin{figure}[h]
    %     \centering
    %         \centering
    %         \includegraphics[trim=30 50 50 50, clip,width=0.24\linewidth, ]{plots/interpolation/vmf_99.png}
    %         \includegraphics[trim=30 50 50 50, clip,width=0.24\linewidth]{plots/interpolation/vmf_60.png}
    %         \includegraphics[trim=30 50 50 50, clip,width=0.24\linewidth]{plots/interpolation/vmf_30.png}
    %         \includegraphics[trim=30 50 50 50, clip,width=0.24\linewidth]{plots/interpolation/vmf_0.png}
    %     \caption{VMF Interpolation: backward at IPF convergence: right to left}
    % \end{figure}

\section{Future Directions}
We introduce novel methodology for generative modeling and interpolation on
compact Riemannian manifolds, which accelerates and generalizes RSGM.  In future
work, we will apply RDSB to more challenging settings such as interpolation in
robotics and for protein modeling, see \citep{jing2022torsional}. From a
theoretical point of view, the restriction to compact spaces allows us to
leverage tools from Optimal Transport \cite{peyre2019computational} to prove the
geometric convergence of RDSB to an approximation of the Schr\"odinger Bridge
and quantify this bias.

\newpage
\bibliography{references}
\bibliographystyle{icml2022}

\end{document}